# Contextual Bandits with Sparse Data in Web setting

Björn H Eriksson
*Institutionen för data- och systemvetenskap, DSV*
*Stockholms universitet*
Stockholm, Sweden
bjheriksson@protonmail.ch

*Abstract*—This paper is a scoping study to identify current methods used in handling sparse data with contextual bandits in web settings. The area is highly current and state of the art methods are identified. The years 2017-2020 are investigated, and 19 method articles are identified, and two review articles. Five categories of methods are described, making it easy to choose how to address sparse data using contextual bandits with a method available for modification in the specific setting of concern. In addition, each method has multiple techniques to choose from for future evaluation. The problem areas are also mentioned that each article covers. An overall updated understanding of sparse data problems using contextual bandits in web settings is given. The identified methods are policy evaluation (off-line and on-line) , hybrid-method, model representation (clusters and deep neural networks), dimensionality reduction, and simulation.

Keywords—*sparse data, contextual bandit, multi-armed bandit, reinforcement learning, web*

## I. Introduction

One of the most straightforward reinforcement learning (RL) agents is called a multi-armed bandit (MAB), referring to a slot machine (a one-armed bandit). Imagine a recommendation engine that can show different products. Each product can be considered an arm, and the agent choose to draw a specific arm when it decides which product to expose to the user. A contextual bandit (CB) adds the context; it takes into account something about the environment. For instance, the user's actions, browsing history, etcetera.

Contextual bandits are common and suitable in web settings for recommendation like problems [1], [2], also for scenarios with sparse data [3]. In fact, sparse data represent one of the main challenges CB are facing. This is clearly articulated in both recent and older articles [4], [1], [5], [6] Current article is a scoping study too enhance the understanding of what methodologies and technics are used for sparse data handling using contextual bandit in a web setting.

Imagine a recommendation engine, and a user browses in. We identify an IP-address and read some historic cookies but no information is present about the preferences in our area related to the user and the product we want to show. The situation might remain even after the user has been exposed to the algorithm because we know still very little individually, due to the few number of interactions.

Li et al. [7] described a contextual bandit approach at Yahoo!, handling the problem around sparse feedback, doing off-line batch training and managed to significantly improve the usefulness of contextual bandits. Others have used dimensionality reduction techniques, they can be applied to compensate for sparse data problems, as was the case in Netflix data, see Koren et al. [8]. This data is interesting because it is well known and at the same time an example of an extremely sparse data set [9]. There are several other approaches to address different aspects of sparse data, some have proposed time-varying contextual information [10] mainly to coop with changing preferences. Hybrid methods with contextual bandit have been developed especially for cold-start problems [11]. This method includes some other technique that adds to the functioning, for instance the simplest is a recommender system that shows five products, three are the most popular products over all and two are the personalized products for the specific user decided by the contextual bandit.

No review study is updated targeting contextual bandits for sparse data to our knowledge. One updated study addressing the data sparsity problem [12] cover several interesting areas however it is not focused on contextual bandits. The study identifies only seven recent articles (2017-2019), and non-of these focus on contextual bandits.

In addition, the ambition of this paper is to find methods to make it easier to choose to address sparse data with a method that is available for modification in a specific setting. The study is hence motivated by this practical perspectives as well as the lack of a current updated overview targeting sparse data and contextual bandits as such.

## II. Methods

This paper is a scoping study and not a systematic review [13]. The aim is to find current technics and methods handlining sparse data with contextual bandits in a web setting. It is important to remember that this problem intersects several problem areas such as the exploration/exploitation dilemma, accuracy/novelty, the cold start problem, and high-dimensional data problems. We aim here to address the sparse data problem in its narrowest sense in a specific context. Moreover, we ask the research question.

*What are the current methods used to handle sparse data when using contextual bandits in a web setting.*

For this study, Google Scholar and EBSCO Discovery Service search engines were consulted. The search was done for "contextual bandit" AND "sparse data" for 2017-2020. The narrow search timeframe is motivated due to the research question concern current use. The narrow search terms can be discussed. However, focusing on the exact intersection of sparse data and contextual bandits, the selection was deemed possible to give a contemporary overview because the likelihood of high grade of the results was highly relevant.

III. RESULTS

In total, 53 results were attained from the search. These were imported to Publish or Perish for better handling and overview. The first nine results were discarded for categorical reasons (non-English, citations, patents, book, duplicate). For the remaining 44 results, a review of the headings was performed. Here seven articles were removed because they focused on areas outside a web setting (retail operations, cancer, warehousing, truck fleet, urban mobility, tutoring).

The remaining 37 papers were reviewed by an abstract study. Three dissertations or dissertation-like results were identified together with nine results deemed too weak for sparse data or too weakly targeting a web setting.

The remaining 26 articles were reviewed in their entirety, excluding two not possible to retrieve due to budget constraints. Three were deemed not sufficiently on target after reviewed. The remaining 21 are part of this study, where 19 represent different technics and methods, and two are review studies.

The reviews was used first to develop a better understanding of the overall situation. The 19 articles were categorized and classified according to a division of five groups of methods where two of these are split on two subgroups each. These are:

Off-line policy evaluation, on-line policy evaluation, hybrid method, model representation using clusters, model representation using Deep Neural Networks, dimensionality reduction and simulations. A short description of each method follows based on the relevant articles after the summarization in tables. First a summarization of the findings, in table one we see the methods used in the studied articles.

*Fig 1: Sparse data methods with specification of solution*

| Method | Specification | Year | Ref. |
|---|---|---|---|
| Policy evaluation | Off-line for correction of policy | 2019 | [11] |
| | Off-line, baseline for safe policy | 2020 | [14] |
| | On-line, joint model | 2020 | [15] |
| | On-line, unbiased estimators | 2018 | [5] |
| Hybrid-method | Personalized and popular items | 2020 | [17] |
| | Interview questions, joint model | 2018 | [16] |
| Model representation | Clusters, knowledge entities | 2019 | [18] |
| | Clusters, decoupled & cat. specific | 2020 | [19] |
| | Clusters, co-clustering | 2019 | [2] |
| | DNN, multiple scheme state rep. | 2020 | [22] |
| | DNN, actor critique | 2018 | [21] |
| | DNN, content and context features | 2019 | [20] |
| | DNN, model uncertainty | 2018 | [25] |
| Dimensionality reduction | Matrix factoring | 2019 | [23] |
| | Joint-model | 2020 | [4] |
| | Random projections | 2019 | [6] |
| Simulation | On-line, joint model with dropouts | 2019 | [24] |
| | Joint-model for simulation | 2017 | [1] |
| | Simulation framework | 2020 | [26] |

Fig. 1. Methods for handling sparse data using contextual bandits.

For context an overview of the articles targeted problem areas are presented in figure two. The method is matched with the chosen problem area. It is outside of the scope of this article to present these problem areas.

*Fig 2: Sparse data problem areas and methods*

| Method | Problem area | Year | Ref. |
|---|---|---|---|
| Policy evaluation | Bias in data sets | 2019 | [11] |
| | Safety in user experience | 2020 | [14] |
| | Counterfactual learning | 2020 | [15] |
| | Biased estimators | 2018 | [5] |
| Hybrid-method | Cold start problem | 2020 | [17] |
| | Cold start problem | 2018 | [16] |
| Model representation | Understanding the context | 2019 | [18] |
| | Sparse data on Amazon | 2020 | [19] |
| | Connect user and content | 2019 | [2] |
| | State representation | 2020 | [22] |
| | Large-scale state representation | 2018 | [21] |
| | Novelty problem | 2019 | [20] |
| | Personalization methods | 2018 | [25] |
| Dimensionality reduction | Handling sparse data | 2019 | [23] |
| | Bias toward dense data structures | 2020 | [4] |
| | High-dimensional sparse data | 2019 | [6] |
| Simulation | Unobserved heterogeneity | 2019 | [24] |
| | Unobserved heterogeneity | 2017 | [1] |
| | Offline-online learning problems | 2020 | [26] |

Fig. 2. Problem areas described in articles about sparse data handling and contextual bandits.

### *Off-line policy evaluation*

One idea is to use off-policy methods to leverage on-line learning in different ways. That is we evaluate on historic data removed from a live situation. With off-line learning, a baseline policy can be developed and used to establish a safe online-learning scenario [14]. This baseline policy will override the on-line learning policy until it outperforms the baseline and is particular suited for sparse data situations with tests on datasets with up to 90% sparsity. In so doing cold start problems can be minimized, helping to decide when to explore on new users. More complex methods can be used, such as when a Deep Neural Network (DNN) was used for policy evaluation, and an off-policy candidate generation occurred together with an off-policy correction to combat bias in the model [11]. That is, the policy was regularly evaluated off-line and possibly updated.

### *On-line policy evaluation*

Counterfactual learning evaluation (we did not get a click for instance) finds that due to sparse rewards, many approaches fail or have weak results. This can be combated with a dual bandit solution in an on-line situation, one example is a Maximum Likelihood Estimation (MLE) and a counterfactual risk minimization (CRM) jointly used for policy evaluation on-line[15]. Another idea is to use what they call a "minmax concave penalized" (MCP) adaptation of a Multi-Armed Bandit. This is reported to outperform benchmark algorithms on sparse data. Naming their algorithm G-MCP-Bandit they show and describe high performance using sparse data while learning unbiased estimators [5].

### *Hybrid method*

A hybrid approach disables the CB in one way or another. Often this is a method of addressing the cold-start problem with the ambition to jumpstart a new user. One solution is straight forward interview questions for new users that can be learned and developed by an RL agent. The new user is asked questions that are used to categorize him or her to let the contextual bandit then take decision based on this categorization [16]. A perhaps more common hybrid approach in recommender settings is to relay on a combination of personalized (using a contextual bandit) and popularity (using a statistical method) based recommendations [17]. This was briefly described in the introduction above.

### Model representation using clusters

Model representation is the part about giving the context meaning. By clustering with unsupervised learning in the data matrix together with evaluating cluster-specific rewards for the policies, a reported and impressive increase in streaming data for music, which was the target, was achieved with 25% in more listening time [2]. One other idea is also to dynamically capture both article and consumer data in "knowledge entities" using a "self-attention mechanism". In this way, an article and a consumer can be contextualized, building up meaning connected to it [18]. Likewise one article proposed decoupled learning rates by using category-specific "meta-priors" by imposing hierarchical structures. In this way, an Empirical Bayes contextual bandit was used in Amazon web shop with good results on sparse data [19]. From a strict accuracy perspective, a reward function for each pair of user and content should be developed, but this is not feasible considering the amount of data and especially the sparsity of data on a single user level, it is simply not possible in a normal sparse data setting. Applying clustering in some sort can solve this problem and overcome the sparsity [2]. In other words, this technique can be viewed as a way to create denser data structures by combing or reducing dimensions in a new space.

### Model representation using Deep Neural Network

Using a Recurrent Neural Network for modeling the combination of content and context features can be a way to address the accuracy and novelty problem [20]. They are rightly used to model a correct representation. Multiple twists on this set-up are possible such as the use of a deep reinforcement learning with an actor-critic approach to model interaction between the user and the system [21]. This can be described as a state representation of the environment using a deep reinforcement learning network [22]. To adapt RL techniques for large-scale situations, a Deep Neural Network can model the interactions between users and items to learn the state representation so as to be able to contribute for sparse data problems [21].

### Dimensionality reduction

Sparse data can be handled with dimensionality reduction. A common technique is called matrix factoring. [23]. A related problem to the number of dimensions is that current recommender systems in production settings tend to prefer products with dense data structures. In all simplicity, this means that the systems tend to recommend high-sales products and not products from the so called long-tail representing more sparse data structures. One effect is that the recommendations might appear boring to the user over time. This can be overcome by incorporating, for example, a gradient boosting decision tree in the bandits used to extract high-level features [4] Others have made similar contributions. For example, for high-dimensional sparse data for personalized assortment decisions, a combining algorithm (OLP-UCB) and a random projection method were successfully described [6].

### Simulations

Here we have collected examples of using simulations for evaluating the algorithms. Proposing a way to account for unobserved heterogeneity by combining Multi-Armed Bandits with hierarchical regression and randomized allocation rules, a great success was achieved in an on-line advertising set-up [1]. This same problem is sometimes referred to as the feedback loop and is a well-recognized problem in all personal recommendation settings. Newly trained algorithms tend to favor items already engaged by the user, similar to the situation with products. This can be addressed, such as in one study with a deep Bayesian bandits algorithm, by combining a deep neural network with "dropouts" where Thompson sampling or upper confidence bound (UCB) is used instead [24]. The combination of a contextual bandit with a deep neural network appears in several studies, such as in a "Deep Uncertainty-Aware Learning (DUAL)" where deep neural networks are used for the model representation, especially uncertainty estimations of the predictions used in this case to navigating the exploration phase [25].

Researchers are aware of the challenges posed by dynamic interactions. Most recommendation algorithms are designed and evaluated off-line. By using a specific simulation framework, one report showed that off-line training data is reasonably predictable of on-line performance and that the simulation soon can reach demising return problems [26].

### Goalsetting

A short note concerns setting the correct goal for the contextual bandit. Sometimes or perhaps often the goal is set for convenience for what is understood as an intermediate goal such as a click. For example in a ads network for customer acquisition one article described that the goal as such had a significant impact on performance, for them, the click-through rate optimization is not the same as conversion optimization. In fact, it performed 12% worse [1] when optimized for click-through instead for conversion (actual acquisition). This might be well to remember and make sure to evaluate on the specific goal if possible.

## IV. DISCUSSION

The results cover the research question, we have made an overview of currently available methods to address sparse data problems for contextual bandits in a web setting. Technics are mentioned but not explained. This might be valuable in the future to evolve on the technics as such. To further evaluate the contribution of these methods, evaluating their impact on the same dataset would help select the correct method. For instance doing standardized tests for the different methods could greatly benefit the selection process. It is also possible to consider a wider search period or more key words. This was outside the scope of this study. Possibly methods used for sparse data handling in other reinforcement learning situations outside the strict contextual bandit situation might be likely, [12] can be consulted for a starting point.

The different techniques described in the articles are only mentioned in this study. However for convince in figure three follows a detailed overview for reference. They are only briefly mentioned as a reference point, it is outside the scope to describe them in this article.

*Fig 3: Sparse data methods with tehniques*

| Meth. | Specific. | Technique | Ref. |
|---|---|---|---|
| Policy Eval. | Off-line for correction of policy | DNN for policy+ off-policy candidate generation + off-policy correction | [11] |
| | Off-line, baseline for safe policy | Use base policy developed off-line, until online gets better results. | [14] |
| | On-line, joint model | MLE + CRM | [15] |
| | On-line, unbiased estimators | G-MCP-Bandit | [5] |
| Hybrid | Personalized and popular items | Combining personalized and popular | [17] |
| | Interview questions, joint model | interview questions for new users managed by an RL agent with Deep Q Network support | [16] |
| Model repr. | Clusters, knowledge entities | both article and consumer are contextualized with the help of self-attention mechanism | [18] |
| | Clusters, decoupled & cat. specific | Decoupled learning rates, using empirical bayes for category specific values | [19] |
| | Clusters, co-clustering | unsupervised learning to create clusters with specific rewards that feeds the policy | [2] |
| | DNN, multiple scheme state rep. | DNN RL | [22] |
| | DNN, actor critique | DNN RL | [21] |
| | DNN, content and context features | DNN RL | [20] |
| | DNN, model uncertainty | Deep Uncertainty-Aware Learning | [25] |
| Dim. red. | Matrix factoring | matrix factoring, selection of proper filtering technique | [23] |
| | Joint-model | CB + gradient boosting DT | [4] |
| | Random projections | OLP-UCB algorithm + random projection method | [6] |
| Simul. | On-line, joint model with dropouts | DNN + dropouts with Thompson sampling or UCB | [24] |
| | Joint-model for simulation | joint model MAB + hierarchical clustering + randomized action rules | [1] |
| | Simulation framework | Used to evaluate if offline data is sufficient for policy evaluation | [26] |

Fig 3. Different techniques described in the articles about sparse data handling and contextual bandits. Possible further study. Provided as is, without comment.

## V. CONCLUSION

This paper have classified 19 articles between the years 2017 – 2020 to identify methods for handling data sparsity using contextual bandits in web settings. We have categorized the articles in 5 different methods with subgroups. We found Policy evaluation (2 off-line and 2 on-line), 2 hybrid-method, model representation (3 clusters and 4 deep neural networks), 3 dimensionality reduction, and 3 simulation, all of them used to handle specific data sparsity problems for contextual bandits in web settings.

The exact method to choose will have to be related to a specific problem and situation. This study offers a convenient overview to identify possible methods to handle sparse data with contextual bandits in a web setting, offering a selection of choices to better handle sparse data, in addition, multiple technics are mentioned for each category of methods.